\documentclass[a4paper]{article}
\usepackage{indentfirst}
\usepackage{subfigure}
\usepackage{textcomp}
\usepackage{graphicx}
\usepackage{amsmath}
\usepackage{amsopn}
\usepackage{amssymb}
\usepackage{url}

\newtheorem{definition}{Definition}

\newtheorem{remark}{Remark}

\begin{document}

\title{A Novel Clustering Algorithm Based on Quantum Games}
\author{Qiang Li, Yan He, Jing-ping Jiang\\ \\
College of Electrical Engineering, Zhejiang University,\\ Hang Zhou,
Zhejiang, 310027, China} \maketitle

\begin{abstract}
The enormous successes have been made by quantum algorithms during
the last decade. In this paper, we combine the quantum game with the
problem of data clustering, and then develop a quantum-game-based
clustering algorithm, in which data points in a dataset are
considered as players who can make decisions and implement quantum
strategies in quantum games. After each round of a quantum game,
each player's expected payoff is calculated. Later, he uses an
link-removing-and-rewiring (LRR) function to change his neighbors
and adjust the strength of links connecting to them in order to
maximize his payoff. Further, algorithms are discussed and analyzed
in two cases of strategies, two payoff matrixes and two LRR
functions. Consequently, the simulation results have demonstrated
that data points in datasets are clustered reasonably and
efficiently, and the clustering algorithms have fast rates of
convergence. Moreover, the comparison with other algorithms also
provides an indication of the effectiveness of the proposed
approach.
\end{abstract}

\maketitle

\section{Introduction}

Quantum computation is an extremely exciting and rapidly growing
field. More recently, an increasing number of researchers with
different backgrounds, ranging from physics, computer sciences and
information theory to mathematics and philosophy, are involved in
researching properties of quantum-based
computation~\cite{Vedral1998}. During the last decade, a series of
significant breakthroughs had been made. One was that in 1994 Peter
Shor surprised the world by proposing a polynomial-time quantum
algorithm for integer factorization \cite{Shor1994}, while in the
classical world the best-known classical factoring algorithm works
in superpolynomial time. Three years later, in 1997, Lov Grover
proved that a quantum computer could search an unsorted database in
the square root of the time \cite{Grover1997}. Meanwhile, Gilles
Brassard et al. combined ideas from Grover's and Shor's quantum
algorithms to propose a quantum counting algorithm
\cite{Brassard1998}.

In recent years, many interests focus on the quantum game theory and
considerable work has been done. For instance, D. A.
Meyer~\cite{Meyer1999} studied the Penny Flip game in the quantum
world firstly. His result showed that if a player was allowed to
implement quantum strategies, he would always defeat his opponent
who played the classical strategies and increase his expected payoff
as well. J. Eisert et al.~\cite{Eisert1999} quantized the Prisoners'
Dilemma and demonstrated that the dilemma could be escaped when both
players resort to quantum strategies. A. P. Flitney et
al.~\cite{Flitney2003} generalized Eisert's result, the miracle
move, i.e., the result of the game would move towards the quantum
player's preferred result, while the other player used classical
strategies. L. Marinatto et al.~\cite{Marinatto2000} investigated
the Battle of the Sexes game in quantum domain. Their result showed
that there existed a unique equilibrium in the game, when the
entangled strategies were allowed. C. F. Lee et al.~\cite{Lee2003}
reported that the quantum game is more efficient than the classical
game, and they found an upper bound for this efficiency. Besides,
some experiments about the quantum games have also been implemented
on different quantum
computers~\cite{Du2002,Prevedel2007,Schmid2009}. For more details
about quantum games, see~\cite{Guo2008}.

Successes achieved by quantum algorithms make us guess that powerful
quantum computers can figure out solutions faster and better than
the best known classical counterparts for certain types of problems.
Furthermore, it is more important that they offer a new way to find
potentially dramatic algorithmic speed-ups. Therefore, we may ask
naturally: can we construct quantum versions of classical algorithms
or present new quantum algorithms to solve the problems in pattern
recognition faster and better on a quantum computer? Following this
idea, some researchers have proposed their novel methods and
demonstrated exciting results
\cite{Horn2001,Sasaki2002,Trugenberger2002a,Schutzhold2003,Ameur2007}.

In addition, data clustering is a main branch of Pattern
Recognition, which is widely used in many fields such as pattern
analysis, data mining, information retrieval and image segmentation.
In these fields, however, there is usually little priori knowledge
available about the data. In response to these restrictions,
clustering methodology come into being which is particularly
suitable for the exploration of interrelationships among data
points. Data clustering is the formal study of algorithms and
methods for grouping or classifying unlabeled data
points~\cite{Jain1999}. In other words, its task is to find the
inherent structure of a given collection of unlabeled data points
and group them into meaningful clusters~\cite{Jain1999}. In this
paper, we attempt to combine the quantum game with the problem of
data clustering in order to establish a novel clustering algorithm
based on quantum games. In our algorithms, unlabeled data points in
a dataset are regarded as players who can make decisions in quantum
games. On a time-varying network formed by players, each player is
permitted to use quantum strategies and plays a $2\times2$ entangled
quantum game against every one of his neighbors respectively. Later,
he applies a link-removing-and-rewiring (LRR) function to remove the
links of neighbors with small payoffs and create new links to
neighbors with higher payoffs at the same time. Furthermore, the
strength of links between a player and his neighbors is different
from one another, which is updated by the Grover iteration. During
quantum games, the structure of network and the strength of links
between players tend toward stability gradually. Finally, if each
player only connects to the neighbor with the highest strength, the
network will naturally divide into several separate parts, each of
which corresponds to a cluster.

The remainder of this paper is organized as follows: Section 2
introduces some important concepts about the quantum computation and
the quantum Prisoners' Dilemma briefly. In Section 3, the algorithms
are established in two cases of strategies, payoff matrices and
link-removing-and-rewiring (LRR) functions, and then they are
elaborated and analyzed. In Section 4, the relationship between the
number of nearest neighbors and the number of clusters is discussed.
Next, the effect of the cost in the SD-like payoff matrix is
analyzed, and the relationship between the total payoffs and the
rates of convergence of algorithms is explained. In Section 5, those
datasets used in the simulations are introduced briefly, and then
results of algorithms are demonstrated. The conclusion is given in
Section 6.

\section{Quantum computation and quantum game}
\subsection{Quantum computation}
The elementary unit of quantum computation is called the qubit,
which is typically a microscopic system, such as an atom, a nuclear
spin, or a polarized photon. In quantum computation, the Boolean
states 0 and 1 are represented by a prescribed pair of normalized
and mutually orthogonal quantum states labeled as
$\{|0\rangle,|1\rangle\}$ to form a 'computational
basis'~\cite{Ekert2001}. Any pure state of the qubit can be written
as a superposition state $\alpha|0\rangle+\beta|1\rangle$ for some
$\alpha$ and $\beta$ satisfying
$|\alpha|^{2}+|\beta|^{2}=1$~\cite{Ekert2001}. A collection of
\textit{n} qubits is called a quantum register of size \textit{n},
which spans a Hilbert space of $2^{n}$ dimensions, so $2^{n}$
mutually orthogonal quantum states can be available.

Quantum state preparations, and any other manipulations on qubits,
have to be performed by unitary operations. A quantum logic gate is
a device which performs a fixed unitary operation on selected qubits
in a fixed period of time, and a quantum circuit is a device
consisting of quantum logic gates whose computational steps are
synchronized in time~\cite{Ekert2001}. The most common quantum gate
is the Hadamard gate, which acts on a qubit in state $|0\rangle$ or
$|1\rangle$ to produce
\begin{equation}
\left\{ \begin{array}{cc} |0\rangle\xrightarrow{H}\frac{1}{\sqrt{2}}|0\rangle+\frac{1}{\sqrt{2}}|1\rangle \vspace{4pt}\\
|1\rangle\xrightarrow{H}\frac{1}{\sqrt{2}}|0\rangle-\frac{1}{\sqrt{2}}|1\rangle\end{array}
\right., H=\frac{1}{\sqrt{2}} \left( \begin{array}{cc}
1 & 1\\
1 & -1
\end{array} \right).
\end{equation}
For more details, see~\cite{Ekert2001,Nielsen2000}.

\subsection{Quantum Prisoners' Dilemma}
The Prisoners' Dilemma (PD), a well-known example in the classical
game theory, is an abstract of many phenomena in the real world and
it has been wildly used in plenty of scientific fields. In this
game, each of two players has two optional strategies, cooperation
(C) and defection (D). Later, he chooses one strategy against the
other's for maximizing his own payoff, so does the other side at the
same time, but both sides do not know the opponent's strategy. As a
result, each player receives a payoff which depends on his selected
strategy, where the payoff matrix under different strategy profiles
is described in Table~\ref{tab:1}.
\begin{table}[htbp]
\caption{Payoff matrix for the Prisoners' Dilemma.}\label{tab:1}
\centering \vspace{4pt}
\begin{tabular}{c|cc}
 & $C=0$ & $D=1$\\
 \hline
$C=0$ & $R=3$ & $S=0$\\ \vspace{4pt} $D=1$ & $T=5$ & $P=1$
\end{tabular}
\end{table}
According to the classical game theory, the strategy profile
(defection, defection) is the unique Nash
Equilibrium~\cite{Nash1950,Nash1951}, but unfortunately it is not
Pareto optimal~\cite{Fudenberg1983}.

\begin{figure}[htbp]
\centering
\includegraphics[width=0.5\textwidth]{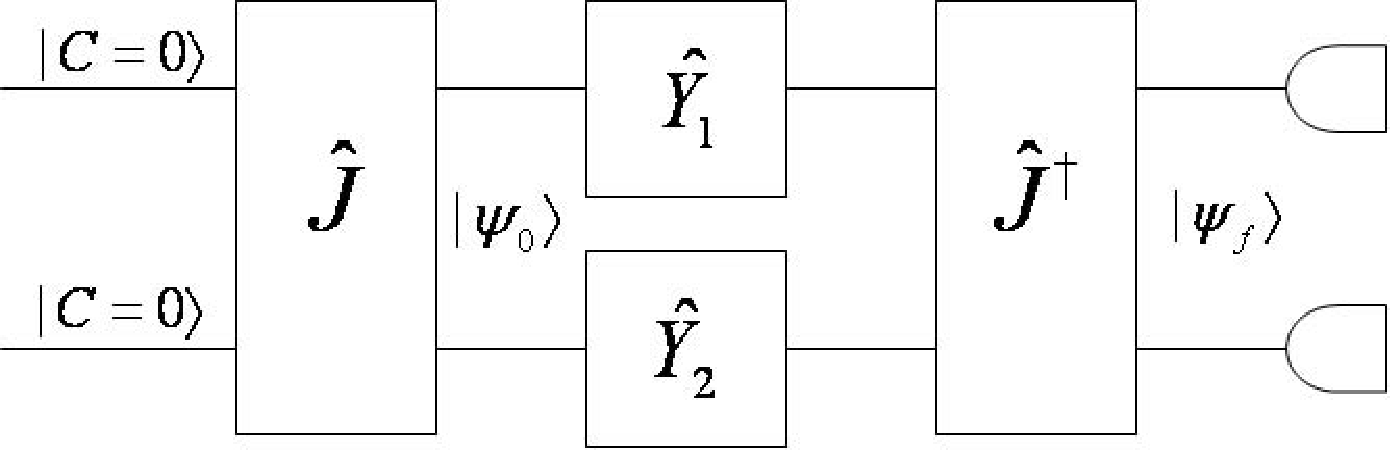}
\caption{The block diagram of the system.}\label{fig:1}
\end{figure}

In the quantum game, however, thanks to the quantum strategies, the
dilemma in the classical game can be escaped in a restricted
strategic space~\cite{Eisert1999}. The physical model of quantum
Prisoners' Dilemma presented by Eisert~\cite{Eisert1999} is shown in
Figure~\ref{fig:1}.

If the possible outcomes of the classical strategies, $C=0$ and
$D=1$, are assigned to two basis vectors
$\{|C=0\rangle,|D=1\rangle\}$ in Hilbert space respectively, then at
any time the state of the game may be represented by a vector in the
space spanned by the basis
$\{|00\rangle,|01\rangle,|10\rangle,|11\rangle\}$~\cite{Eisert1999}.
Assume the initial state of the game is
$|\psi_{0}\rangle=\hat{J}~|00\rangle$, where $\hat{J}$ is an
entangling operator which is known to both players. For a two-player
game with two pure strategies, the general form of $\hat{J}$ may be
written as~\cite{Benjamin2001,Du2001}
\begin{equation}
\hat{J}(\gamma)=exp(i\frac{\gamma}{2}\sigma_{x}^{\otimes2})=I^{\otimes2}cos\frac{\gamma}{2}+i\sigma_{x}^{\otimes2}sin\frac{\gamma}{2}
\end{equation}
where $\gamma\in[0,\pi/2]$ is a measure of entanglement of a game.
When $\gamma=\pi/2$, there is a maximally entangled game, in which the entangling operator takes form
\begin{equation}
\hat{J}(\gamma)=\frac{1}{\sqrt{2}}~(I^{\otimes2}+i\sigma_{x}^{\otimes2}).
\end{equation}
Next, each player chooses a unitary operator
$\hat{Y_{1}}(\hat{Y_{2}})$ from the strategy space $S_{1}(S_{2})$
and operates it on the qubit that belongs to him, which makes the
game in a state $(\hat{Y_{1}}\otimes\hat{Y_{2}})\hat{J}|00\rangle$.
Specifically, the unitary operators $\hat{C}$ and $\hat{D}$ that correspond to the strategies, cooperation and defection, are given below~\cite{Eisert1999}
\begin{equation}
\hat{C}=\begin{pmatrix}1 & 0 \\ 0 & 1\end{pmatrix},\hspace{4pt}
\hat{D}=\begin{pmatrix}0 & 1 \\ -1 & 0\end{pmatrix}.
\end{equation}

In the end, before a projective measurement in the basis
$\{|0\rangle,|1\rangle\}$ is carried out, the final state is
\begin{equation}
|\psi_{f}\rangle=\hat{J}^{\dag}(\hat{Y_{1}}\otimes\hat{Y_{2}})\hat{J}~|00\rangle.
\end{equation}
Thus, the player's expected payoff is written as
\begin{equation}
z=R|\langle\psi_{f}|00\rangle|^{2}+S|\langle\psi_{f}|01\rangle|^{2}+T|\langle\psi_{f}|10\rangle|^{2}+P|\langle\psi_{f}|11\rangle|^{2}.
\end{equation}
For more details, see~\cite{Eisert1999,Du2003}.

\section{Algorithm}
In the section, we will combine the model of the quantum game with
the problem of data clustering, and then establish clustering
algorithms based on quantum games. Assume an unlabeled dataset
$\textbf{\textit{X}}=\{\textbf{\textit{X}}_{1},\textbf{\textit{X}}_{2},\cdots,\textbf{\textit{X}}_{N}\}$,which
are distributed in a \textit{m}-dimensional metric space. Each data
point in the dataset is considered as a player in quantum games who
can make decisions and always hope to maximize his payoff. In the
metric space, there is a distance function $d:\textbf{\textit{X}}
\times \textbf{\textit{X}} \longrightarrow \mathbb{R}$, satisfying
the closer the two players are, the smaller the output is. Based on
the distance function, a \textit{k} nearest neighbors (knn) network
as a weighted and directed network,
$G_{0}(\textbf{\textit{X}},E_{0},d)$, may be created among data
points by adding \textit{k} edges directed toward its \textit{k}
nearest neighbors for each player.

\begin{definition}
If there is a set $\textbf{\textit{X}}$ with $N$ players,
$\textbf{\textit{X}}=\{\textbf{\textit{X}}_{1},\textbf{\textit{X}}_{2},\cdots,\textbf{\textit{X}}_{N}\}$,
the weighted and directed knn network,
$G_{0}(\textbf{\textit{X}},E_{0},d)$, is created as below.
\begin{equation}
\left\{\begin{array}{ll}\textbf{\textit{X}}=\big\{\textbf{\textit{X}}_{i},i=1,2,\cdots,N\big\}\vspace{4pt}\\
E_{0}=\bigcup^{N}_{i=1}E_{0}(i)\vspace{4pt}\\
E_{0}(i)=\Big\{e_{0}\big(\textbf{\textit{X}}_{i},\textbf{\textit{X}}_{j}\big)\mid
j\in\Gamma_{0}(i)\Big\}\vspace{4pt}\\
\Gamma_{0}(i)=\bigg\{j \Big| j=\underset{\textbf{\textit{X}}_{h}\in
\textbf{\textit{X}}}{argmink}\bigg(\Big\{d(\textbf{\textit{X}}_{i},\textbf{\textit{X}}_{h}),\textbf{\textit{X}}_{h}\in\textbf{\textit{X}}\Big\}\bigg)\bigg\}
\end{array} \right.
\end{equation}
Here, each player in the set $\textbf{\textit{X}}$ corresponds to a
vertex in the network $G_{0}(\textbf{\textit{X}},E_{0},d)$; $E_{0}$
is a link set and a link in the network represents certain
relationship between a pair of players; the distances denote the
weights over links; the function, $argmink(\cdot)$, is to find
\textit{k} nearest neighbors of a player which construct a neighbor
set, $\Gamma_{0}(i)$; the subscript '0' is the initial time step.
\end{definition}

It is worth noting that the strength of links between a player
$\textbf{\textit{X}}_{i}$ and his \textit{k} nearest neighbors
represented by
$\rho_{t-1}(\textbf{\textit{X}}_{i},\textbf{\textit{X}}_{j}),j\in\Gamma_{t-1}(i)(t\geq1)$
is time-varying, whose initial values is calculated by
\begin{equation}
\rho_{0}(\textbf{\textit{X}}_{i},\textbf{\textit{X}}_{j})=\bigg\{
\begin{array}{lll}
1/|\Gamma_{0}(i)|=1/k, & \textrm{ $j\in\Gamma_{0}(i)$}\vspace{4pt}\\
0, & \textrm{ otherwise}
\end{array}
\end{equation}
where the symbol $|\cdot|$ denotes the cardinality of a set.

After the initial connections are constructed among players (data
points), on this weighted and directed knn network, a quantum game
can be defined as following.
\begin{definition}
A quantum game $\Omega=\{\textbf{\textit{X}},G_{t},S,Z_{t}\}$ on a
network $G_{t}$ is a 4-tuple: $\textbf{\textit{X}}$ is a set of
players; $G_{t}$ represents the connections among players;
$S=\{s(i),i=1,2,\cdots,N\}$ represents a set of players' strategies
including the full range of quantum strategies;
$Z_{t}=\{z_{t}(i),i=1,2,\cdots,N\}$ represents a set of players'
expected payoffs. Here, the variable $t$ denotes the time step (the
number of iterations). In each round, players choose theirs
strategies simultaneously, and each player can only observe its
neighbors' payoffs, but does not know the strategy profiles of all
other players in $\textbf{\textit{X}}$.
\end{definition}

\subsection{Cases of quantum strategies and payoff matrices}
At first, each player selects a strategy from his strategy set, and
then plays a $2\times2$ entangled quantum game against one of his
\textit{k} neighbors respectively. In the classical $2\times2$ game,
such as the Prisoners' Dilemma, usually there are only two pure
strategies, cooperation and defection, but in the quantum game, one
can design different unitary operators as strategies, i.e., the
strategy set \textit{S} may be identified with some subset of the
group of $2\times2$ unitary matrices~\cite{Eisert1999}. Here, for
the purpose of clustering and simplifying computation, the strategy
set of a player $\textbf{\textit{X}}_{i}$ is restricted in a set
$S_{1}=\{\hat{H},\hat{D}\}$ or
$S_{2}=\{\hat{F}_{t-1}(i,j),\hat{D}\}$, and then two cases of
strategy sets are described respectively.

Case 1:

In this case, a player and his opponent can apply strategies in
$S_{1}=\{\hat{H},\hat{D}\}$. When a player $\textbf{\textit{X}}_{i}$
use the Hadamard matrix $\hat{H}$ as a strategy, his opponent
(neighbor) $\textbf{\textit{X}}_{j}$ has two optional strategies
$\{\hat{H},\hat{D}\}$, but which strategy is chosen is dependent on
the strength of the link between them, as is a rule of the
clustering algorithm. If the strength
$\rho_{t-1}(\textbf{\textit{X}}_{j},\textbf{\textit{X}}_{i})$ equals
to zero, i.e., there is no link directed from the player
$\textbf{\textit{X}}_{j}$ to the player $\textbf{\textit{X}}_{i}$,
then the player $\textbf{\textit{X}}_{j}$ will apply the strategy
'Defection' ($\hat{D}$). Alternatively, if
$\rho_{t-1}(\textbf{\textit{X}}_{j},\textbf{\textit{X}}_{i})>0$,
namely mutual connections between them, the player
$\textbf{\textit{X}}_{j}$ implements the strategy $\hat{H}$. If the
initial state of the game is $|\psi_{0}\rangle=\hat{J}|00\rangle$,
by applying the model of the quantum game the final state of the
game is
\begin{equation}
|\psi_{f,j}\rangle=\Bigg\{\begin{array}{ll}
\frac{1}{2}(|00\rangle-i|01\rangle-i|10\rangle+|11\rangle), & \textrm{ $\hat{H}\otimes \hat{H}$}\vspace{4pt}\\
\frac{1}{\sqrt{2}}(i|00\rangle-|01\rangle), & \textrm{
$\hat{H}\otimes \hat{D}$}
\end{array}
\end{equation}

Case 2:

The strategy set $S_{2}=\{\hat{F}_{t-1}(i,j),\hat{D}\}$ is adopted
in the case. The strategy $\hat{F}_{t-1}(i,j)$,
\begin{equation*}
\hat{F}_{t-1}(i,j)=\begin{pmatrix}\sqrt{\rho_{t-1}(\textbf{\textit{X}}_{i},\textbf{\textit{X}}_{j})} & \sqrt{1-\rho_{t-1}(\textbf{\textit{X}}_{i},\textbf{\textit{X}}_{j})}\\
\sqrt{1-\rho_{t-1}(\textbf{\textit{X}}_{i},\textbf{\textit{X}}_{j})}
&
-\sqrt{\rho_{t-1}(\textbf{\textit{X}}_{i},\textbf{\textit{X}}_{j})}\end{pmatrix},
\end{equation*}
is a general form of Hadamard matrix $\hat{H}$ whose elements are
associated with the strength of links
$\rho_{t-1}(\textbf{\textit{X}}_{i},\textbf{\textit{X}}_{j})$. When
$\rho_{t-1}(\textbf{\textit{X}}_{i},\textbf{\textit{X}}_{j})$=0.5,
the strategy $\hat{F}_{t-1}(i,j)$ recovers the strategy $\hat{H}$.
Similarly, the neighbor $\textbf{\textit{X}}_{j}$, when
$\rho_{t-1}(\textbf{\textit{X}}_{j},\textbf{\textit{X}}_{i})=0$,
applies the strategy 'Defection' ($\hat{D}$), while using the
strategy $\hat{F}_{t-1}(i,j)$ when
$\rho_{t-1}(\textbf{\textit{X}}_{j},\textbf{\textit{X}}_{i})>0$. If
the initial state of the game is
$|\psi_{0}\rangle=\hat{J}|00\rangle$, after their moves, the final
state of the game is
\begin{equation}
|\psi_{f,j}\rangle=\left\{\begin{array}{ll}
\sqrt{\rho_{1}\rho_{2}}|00\rangle-i\sqrt{\rho_{2}(1-\rho_{1})}|01\rangle\\
-i\sqrt{\rho_{1}(1-\rho_{2})}|10\rangle+\sqrt{(1-\rho_{1})(1-\rho_{2})}|11\rangle, & \textrm{ $\hat{F}_{t-1}\otimes \hat{F}_{t-1}$}\vspace{4pt}\\
i\sqrt{1-\rho_{1}}|00\rangle-\sqrt{\rho_{1}}|01\rangle, & \textrm{
$\hat{F}_{t-1}\otimes \hat{D}$}
\end{array}\right..
\end{equation}

According to the payoff matrix, the player's expected payoff can be
computed by
\begin{equation}\label{eq:11}
\begin{array}{ll}
z_{t-1}(i)=\sum_{j\in\Gamma_{t-1}(i)}z_{t-1}(\textbf{\textit{X}}_{i},\textbf{\textit{X}}_{j})\vspace{4pt}\\
=\sum_{j\in\Gamma_{t-1}(i)}R|\langle\psi_{f,j}|00\rangle|^{2}+S|\langle\psi_{f,j}|01\rangle|^{2}
+T|\langle\psi_{f,j}|10\rangle|^{2}+P|\langle\psi_{f,j}|11\rangle|^{2}.
\end{array}
\end{equation}
In practice, the payoff matrix takes PD-like or Snowdrift (SD)-like
form, described in Table~\ref{tab:2} and \ref{tab:3}.
\begin{table}[htbp]
\caption{PD-like payoff matrix.}\label{tab:2} \centering
\vspace{4pt}
\begin{tabular}{c|cc}
 & $C=0$ & $D=1$\\
 \hline
$C=0$ &
$R=0.6\omega_{t-1}(\textbf{\textit{X}}_{i},\textbf{\textit{X}}_{j})$
&
$S=0.01\omega_{t-1}(\textbf{\textit{X}}_{i},\textbf{\textit{X}}_{j})$\\
\vspace{4pt} $D=1$ &
$T=\omega_{t-1}(\textbf{\textit{X}}_{i},\textbf{\textit{X}}_{j})$ &
$P=0.2\omega_{t-1}(\textbf{\textit{X}}_{i},\textbf{\textit{X}}_{j})$
\end{tabular}
\end{table}
\begin{table}[htbp]
\caption{SD-like payoff matrix.}\label{tab:3} \centering
\vspace{4pt}
\begin{tabular}{c|cc}
 & $C=0$ & $D=1$\\
 \hline
$C=0$ &
$R=\omega_{t-1}(\textbf{\textit{X}}_{i},\textbf{\textit{X}}_{j})-\frac{c}{2}$
&
$S=\omega_{t-1}(\textbf{\textit{X}}_{i},\textbf{\textit{X}}_{j})-c$\\
\vspace{4pt} $D=1$ &
$T=\omega_{t-1}(\textbf{\textit{X}}_{i},\textbf{\textit{X}}_{j})$ &
$P=0.01\omega_{t-1}(\textbf{\textit{X}}_{i},\textbf{\textit{X}}_{j})$
\end{tabular}
\end{table}
In the PD-like payoff matrix, the following inequalities holds:
$T>R>P>S$ and $2R>T+S$~\cite{Hauert2004}. To avoid a case that a player's expected
payoff is zero, the variable $S$ in Table~\ref{tab:2} takes a small
value instead of zero in Table~\ref{tab:1}. In addition, the
Snowdrift game assumes that two drivers are blocked by a snowdrift,
each of whom is in either side of the snowdrift. If they want to go
back home, one of them or both must shovel a path through the
snowdrift. So, there exists a cost $c$ in the SD-like payoff matrix
and the following inequality holds: $T>R>S>P$~\cite{Hauert2004},
where
$c=\beta\cdot\omega_{t-1}(\textbf{\textit{X}}_{i},\textbf{\textit{X}}_{j})$
and $\beta$ is a proportional factor. Besides, the variable
$\omega_{t-1}(\textbf{\textit{X}}_{i},\textbf{\textit{X}}_{j})$ in
two payoff matrices is calculated by the formulation below.
\begin{equation}
\omega_{t-1}(\textbf{\textit{X}}_{i},\textbf{\textit{X}}_{j})=\rho_{t-1}(\textbf{\textit{X}}_{i},\textbf{\textit{X}}_{j})\times
Deg_{t-1}(\textbf{\textit{X}}_{j})/d(\textbf{\textit{X}}_{i},\textbf{\textit{X}}_{j})
\end{equation}

\subsection{Design of LRR functions}
When all players' payoffs have been computed, each player will
observe his neighbors' payoffs, and apply a
link-removing-and-rewiring (LRR) function $L_{i}(\cdot)$ to change
his links. A LRR function is defined as below.

\begin{definition}
The LRR function $L_{i}(\cdot)$ is a function of payoffs, whose
output is a set with \textit{k} elements, namely an updated neighbor
set $\Gamma_{t}(i)$ of a player $\textbf{\textit{X}}_{i}$. It is
given as below.
\begin{equation}\label{eq:13}
\begin{array}{lll}
\Gamma_{t}(i)=L_{i}\big(\hat{z}_{t-1}(i)\big)=\underset{j\in\Gamma_{t-1}(i)\bigcup\Upsilon_{t-1}(i)}{argmaxk}\Big(\big\{z_{t-1}(j),j\in\Gamma_{t-1}(i)\bigcup\Upsilon_{t-1}(i)\big\}\Big)\vspace{4pt}\\
\hat{z}_{t-1}(i)=\Big\{z_{t-1}(j),j\in\Gamma_{t-1}(i)\bigcup\Upsilon_{t-1}(i)\Big\},\Upsilon_{t-1}(i)=\bigcup_{j\in\Gamma^{+}_{t-1}(i)}\Gamma_{t-1}(j)\vspace{4pt}\\
\Gamma^{+}_{t-1}(i)=\Big\{j|z_{t-1}(j)\geq\theta_{t-1}(i),j\in\Gamma_{t-1}(i)\Big\},\Gamma^{-}_{t-1}(i)=\Gamma_{t-1}(i)\backslash\Gamma^{+}_{t-1}(i)
\end{array}
\end{equation}
where $\theta_{t-1}(i)$ is a payoff threshold, $\Upsilon_{t-1}(i)$
is called an extended neighbor set, and the function
$argmaxk(\cdot)$ is to find \textit{k} neighbors with the first
\textit{k} largest payoffs in the set
$\Gamma_{t-1}(i)\bigcup\Upsilon_{t-1}(i)$.
\end{definition}

Here, two LRR functions $L^{1}_{i}(\cdot)$ and $L^{2}_{i}(\cdot)$
are designed. The function $L^{1}_{i}(\cdot)$ always observes an
extended neighbor set formed by half neighbors of a data point
$\textbf{\textit{X}}_{i}$,
$\alpha=\lceil~0.5\times|\Gamma_{t-1}(i)|~\rceil$, where the symbol
$\lceil\cdot\rceil$ is to take an integer part of a number
satisfying the integer part is no larger than the number. Next, the
payoff threshold $\theta^{1}_{t-1}(i)$ is set by
$\theta^{1}_{t-1}(i)=find^{\alpha}(\{z_{t-1}(i),j\in\Gamma_{t-1}(i)\})$,
where the function $find^{\alpha}(\cdot)$ is to find the
\textit{$\alpha$}-th largest payoff in a set that contains all
neighbors' payoffs of the data point. When the LRR function
$L^{1}_{i}(\cdot)$ is applied, the links connecting to the neighbors
with small payoffs are removed and meanwhile new links are created
between the data point and found players with higher payoffs. Hence,
according to Eq.(\ref{eq:13}), the new neighbor set is
$\Gamma_{t}(i)=L^{1}_{i}(\hat{z}_{t-1}(i))$.

Unlike the LRR function $L^{1}_{i}(\cdot)$, the LRR function
$L^{2}_{i}(\cdot)$ adjusts the number of neighbors dynamically
instead of the constant number of neighbors in $L^{1}_{i}(\cdot)$.
Therefore, the payoff threshold $\theta^{2}_{t-1}(i)$ takes the
average of neighbors' payoffs,
$\theta^{2}_{t-1}(i)=\sum_{j\in\Gamma_{t-1}(i)}z_{t-1}(i)/|\Gamma_{t-1}(i)|$.
Next, the set $\Gamma^{+}_{t-1}(i)$ is formed according to
Eq.(\ref{eq:13}),
$\Gamma^{+}_{t-1}(i)=\{j|z_{t-1}(j)\geq\theta^{2}_{t-1}(i),j\in\Gamma_{t-1}(i)\}$,
and then the new neighbor set is achieved by means of the LRR
function $L^{2}_{i}(\cdot)$,
$\Gamma_{t}(i)=L^{2}_{i}(\hat{z}_{t-1}(i))$. In the case, when the
payoffs of all neighbors are equal to the payoff threshold
$\theta^{2}_{t-1}(i)$, the output of the LRR function is
$\Gamma_{t}(i)=\Gamma_{t-1}(i)$. This may be viewed as
self-protective behavior for avoiding a payoff loss due to no enough
information acquired.

The LRR function $L_{i}(\cdot)$ expands the view of a player
$\textbf{\textit{X}}_{i}$, i.e., it makes him observe payoffs of
players in the extended neighbor set, which provides a chance to
find players with higher payoffs around him. If no players with
higher payoffs are found in the extended neighbor set, namely
$min(\{z_{t-1}(j),j\in\Gamma_{t-1}(i)\})\geq
max(\{z_{t-1}(h),h\in\Upsilon_{t-1}(i)\})$, then the output of the
LRR function is $\Gamma_{t}(i)=\Gamma_{t-1}(i)$. Otherwise, players
with small payoffs will be removed together with the corresponding
links from the neighbor set and link set, and replaced by some found
players with higher payoff. This process is repeated till the
payoffs of unlinked players in the extended neighbor set are no
larger than those of linked neighbors. Since the links among
players, namely the link set $E_{0}$ in the network
$G_{0}(\textbf{\textit{X}},E_{0},d)$, are changed by the LRR
function, the network $G_{t}(\textbf{\textit{X}},E_{t},d)$ has begun
to evolve over time, when  $t\geq1$.
\begin{equation}
G_{t}(\textbf{\textit{X}},E_{t},d)=\left\{\begin{array}{ll}
\textbf{\textit{X}}(t)=\Big\{\textbf{\textit{X}}_{i}(t),i=1,2,\cdots,N\Big\}\vspace{4pt}\\
\Gamma_{t}(i)=L_{i}(\hat{z}_{t-1}(i))\vspace{4pt}\\
E_{t}=\bigcup^{N}_{i=1}E_{t}(i)\vspace{4pt}\\
E_{t}(i)=\Big\{e_{t}\big(\textbf{\textit{X}}_{i},\textbf{\textit{X}}_{j}\big)\mid
j\in\Gamma_{t}(i)\Big\}\end{array} \right.\vspace{4pt}\\
\end{equation}

\subsection{Strength of links updating}
After the LRR function is applied, the strength of links of players
needs to be formed and adjusted. The new strength of links of a
player $\textbf{\textit{X}}_{i}\in\textbf{\textit{X}}$ is formed by
means of the below formulation.
\begin{equation}
\rho_{t}(\textbf{\textit{X}}_{i},\textbf{\textit{X}}_{j})=\left\{\begin{array}{ll}\frac{\sum_{h\in\Gamma_{t-1}(i)\backslash\{\Gamma_{t-1}(i)\bigcap\Gamma_{t}(i)\}}\rho_{t-1}(\textbf{\textit{X}}_{i},\textbf{\textit{X}}_{h})}{\Big|\Gamma_{t}(i)\backslash
\big\{\Gamma_{t-1}(i)\bigcap\Gamma_{t}(i)\big\}\Big|}
& \textrm{ $j\in\Gamma_{t}(i)\backslash\big\{\Gamma_{t-1}(i)\bigcap\Gamma_{t}(i)\big\}$}\vspace{4pt}\\
\rho_{t-1}(\textbf{\textit{X}}_{i},\textbf{\textit{X}}_{j}) &
\textrm{ otherwise}
\end{array} \right.
\end{equation}
Then, the player adjusts the strength of links as follows. First, he
finds a neighbor $\textbf{\textit{X}}_{m}$ with maximal payoff in
his neighbor set,
\begin{equation}
m=\underset{j\in\Gamma_{t}(i)}{argmax}\Big(\big\{z_{t-1}(j),j\in\Gamma_{t}(i)\big\}\Big)
\end{equation}
Next, the strength of link
$\rho_{t}(\textbf{\textit{X}}_{i},\textbf{\textit{X}}_{j}),
j\in\Gamma_{t}(i)$ is taken its square root and the player
$\textbf{\textit{X}}_{m}$'s strength of the link becomes negative,
\begin{equation}
\left\{\begin{array}{ll}\big\{\sqrt{\rho_{t}(\textbf{\textit{X}}_{i},\textbf{\textit{X}}_{j})},j\in\Gamma_{t}(i)\big\}\vspace{4pt}\\
\sqrt{\rho_{t}(\textbf{\textit{X}}_{i},\textbf{\textit{X}}_{m})}=-\sqrt{\rho_{t}(\textbf{\textit{X}}_{i},\textbf{\textit{X}}_{m})},m\in\Gamma_{t}(i)
\end{array} \right.
\end{equation}
Further, let
$Ave_{t}(i)=(\sum_{j\in\Gamma_{t}(i)}\sqrt{\rho_{t}(\textbf{\textit{X}}_{i},\textbf{\textit{X}}_{m})})/|\Gamma_{t}(i)|$,
thus, the updated strength of link is
\begin{equation}
\rho_{t}(\textbf{\textit{X}}_{i},\textbf{\textit{X}}_{j})=\Big(2\times
Ave_{t}(i)-\sqrt{\rho_{t}(\textbf{\textit{X}}_{i},\textbf{\textit{X}}_{j})}\Big)^{2},j\in\Gamma_{t}(i).
\end{equation}

The above-mentioned method is a variant of the Grover iteration $G$
in the quantum search algorithm \cite{Grover1997}, a well-known
algorithm in quantum computation, which is a way to adjust the
probability amplitude of each term in a superposition state. By
adjustment, the probability amplitude of the wanted is increased,
while the others are reduced. This whole process may be regarded as
the \emph{inversion about average} operation \cite{Grover1997}. For
our case, each strength of link is taken its square root first, and
then the average $Ave_{t}(i)$ of square roots is computed. Finally,
all values are inverted about the average. There are three main
reasons that we select the modified Grover iteration to update the
strength of links: (a) the sum of updated strength of links retains
one,
$\sum_{j\in\Gamma_{t}(i)}\rho_{t}(\textbf{\textit{X}}_{i},\textbf{\textit{X}}_{j})=1$,
(b) certain strength of links between a player and his neighbors is
much larger than the others,
$\rho_{t}(\textbf{\textit{X}}_{i},\textbf{\textit{X}}_{j})\gg
\rho_{t}(\textbf{\textit{X}}_{i},\textbf{\textit{X}}_{h}),h\in\Gamma_{t}(i)\backslash
j$, after the strength of links is updated, and (c) it helps
players' payoffs to follow a power law distribution, in which only a
few players' payoffs are far larger than others' in the end of
iterations.

Besides, the process that all players adjust their strength is order
irrelevant, because the strength of links of all players is updated
synchronously, and the network is a directed network, so the
strength cannot be overridden by other players. When the strength of
links of each player has been updated, an iteration is completed. In
conclusion, when $t\geq1$ , the structure of network representing
connections among players begins to evolve over time.

\section{Discussions}
In the section, firstly, the relationship between the number of
nearest neighbors and the number of clusters is discussed, and then
how the cost in the SD-like payoff matrix influences the results is
analyzed, which provides a way to choose the proportional factor.
Finally, the total payoffs based on two different payoff matrices
are compared and the relationship between the total payoffs and the
rates of convergence of algorithms is explained.

\subsection{Number of nearest neighbors vs. number of clusters}
The number \textit{k} of nearest neighbors represents the number of
neighbors to which a data point (player)
$\textbf{\textit{X}}_{i}\in\textbf{\textit{X}}$ connects. For a
dataset, the number \textit{k} of nearest neighbors determines the
number of clusters in part. Generally speaking, the number of
clusters decreases inversely with the number \textit{k} of nearest
neighbors. For example, when the number \textit{k} of nearest
neighbors is small, it is indicated that the player
$\textbf{\textit{X}}_{i}$ connects to only a few neighbors. At this
time only those not-too-distance neighbors can be observed by the
LRR function, which means that the elements in the union of the
extended neighbor set $\Upsilon_{t-1}(i)$ and the neighbor set
$\Gamma_{t-1}(i)$ are few. Therefore, when the evolution of the
network formed by players is ended, many small clusters are
established among data points. On the other hand, a big number
\textit{k} of nearest neighbors provides more neighbors for each
player, as specifies that the cardinality of the union is larger
than that when a small \textit{k} is taken. This means that more
neighbors can be observed and explored by the LRR function, so that
big clusters containing more data points are formed.

\begin{figure}[htbp]
\centering \subfigure[$ k=9$]{ \label{fig2:subfig:a}
\includegraphics[width=0.3\textwidth]{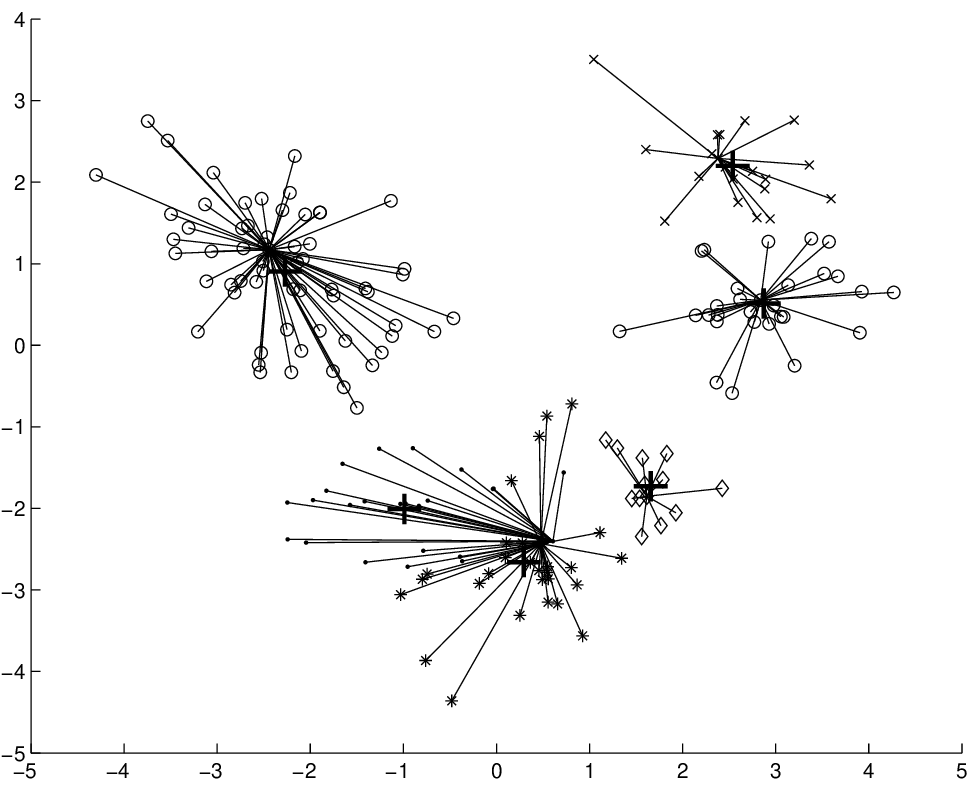}}
\subfigure[$ k=12$]{ \label{fig2:subfig:b}
\includegraphics[width=0.3\textwidth]{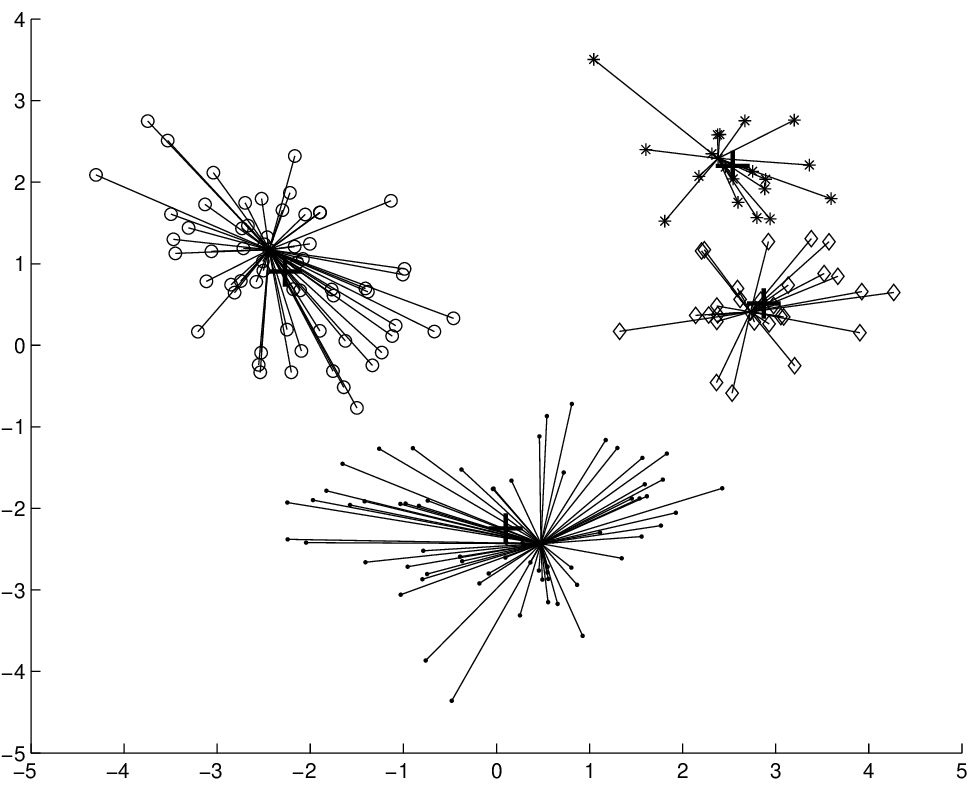}}
\subfigure[$ k=16$]{ \label{fig2:subfig:c}
\includegraphics[width=0.3\textwidth]{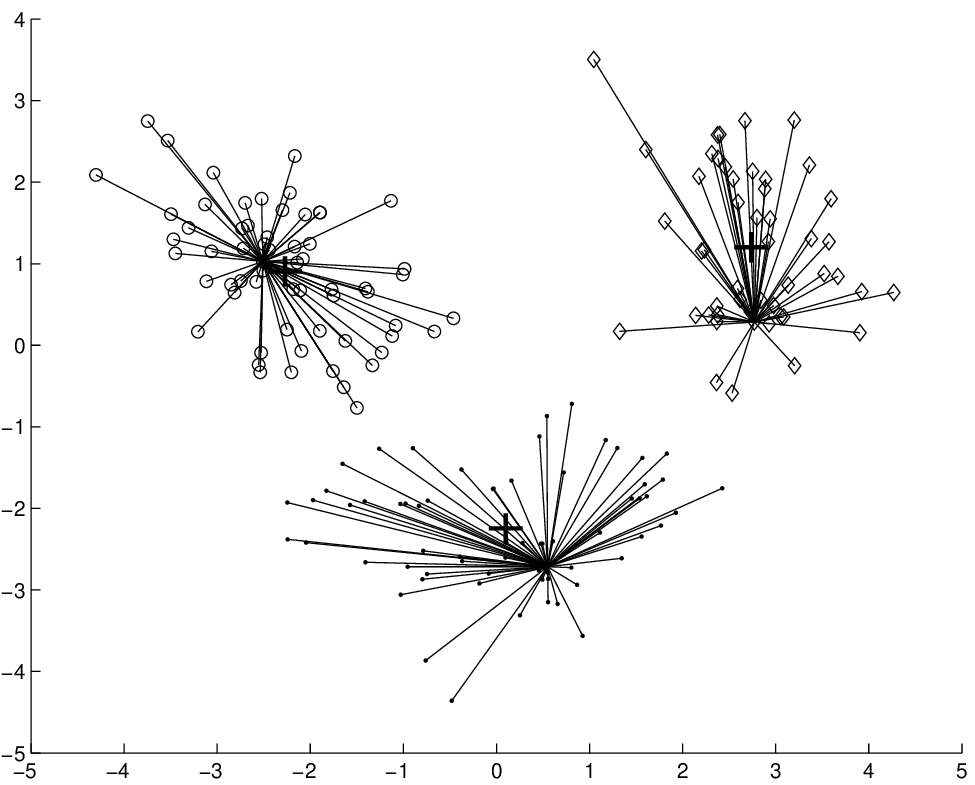}}\\
\caption{The number of nearest neighbors vs. number of clusters. (a)
six clusters are obtained, when the number of nearest neighbors is
$k=9$. (b) four clusters, when $k=12$. (c) three clusters, when
$k=16$}\label{fig:2}
\end{figure}

For a dataset, the clustering results at the different number
\textit{k} of nearest neighbors have been illustrated in
Fig.~\ref{fig:2}, in which each data point only connects to one of
its neighbors who has the largest strength of link, and clusters are
represented by different signs. As is shown in Fig.~\ref{fig:2}, it
can be found that only a few data points receive considerable links,
whereas most of data points have only one link. This implies that
when the structure of network tends to stability, the network, if
only the links with the largest strength are remained, is
characterized by the scale-free network \cite{Barabasi2003}, i.e.,
winner takes all. Besides, in Fig.~\ref{fig:2}(a), six clusters are
obtained by the clustering algorithm, when $k=9$. As the number
\textit{k} of nearest neighbors rises, four clusters are obtained
when $k=12$, three clusters when $k=16$. So, if the exact number of
clusters is not known in advance, different numbers of clusters may
be achieved by adjusting the number of nearest neighbors in
practice.

\subsection{Effect of the cost $c$ in the SD-like payoff matrix}
\begin{figure}[htbp]
\centering \subfigure[]{ \label{fig3:subfig:a}
\includegraphics[width=0.45\textwidth]{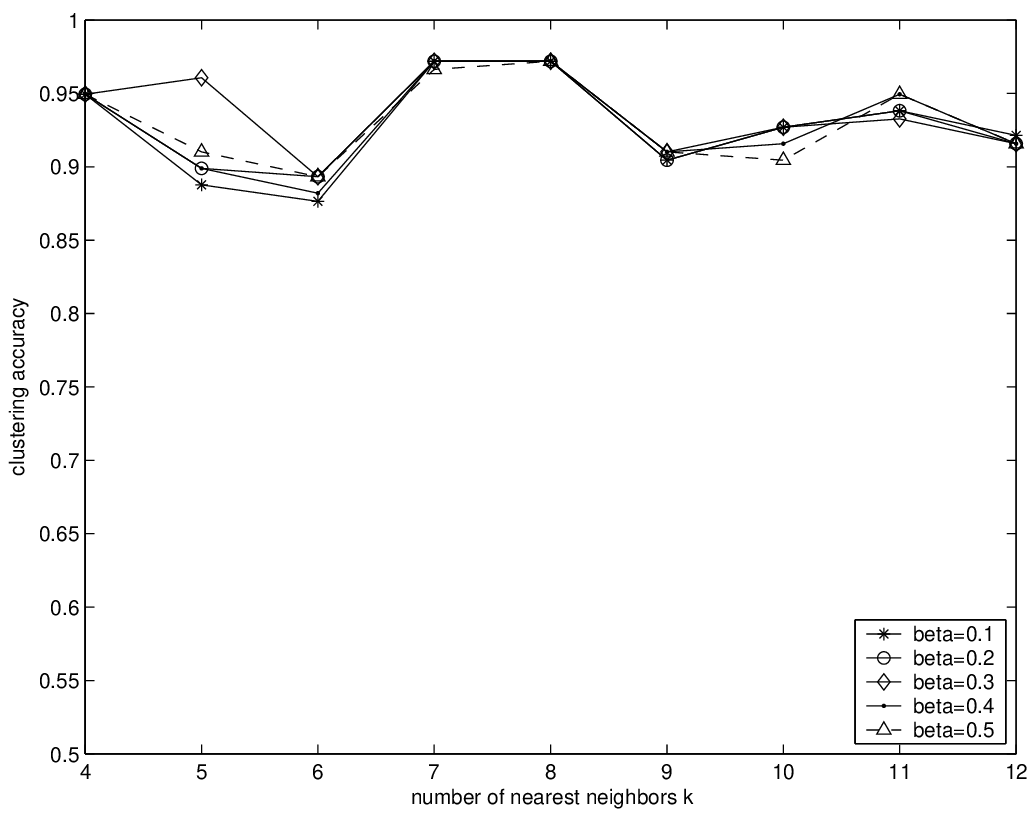}}
\subfigure[]{ \label{fig3:subfig:b}
\includegraphics[width=0.45\textwidth]{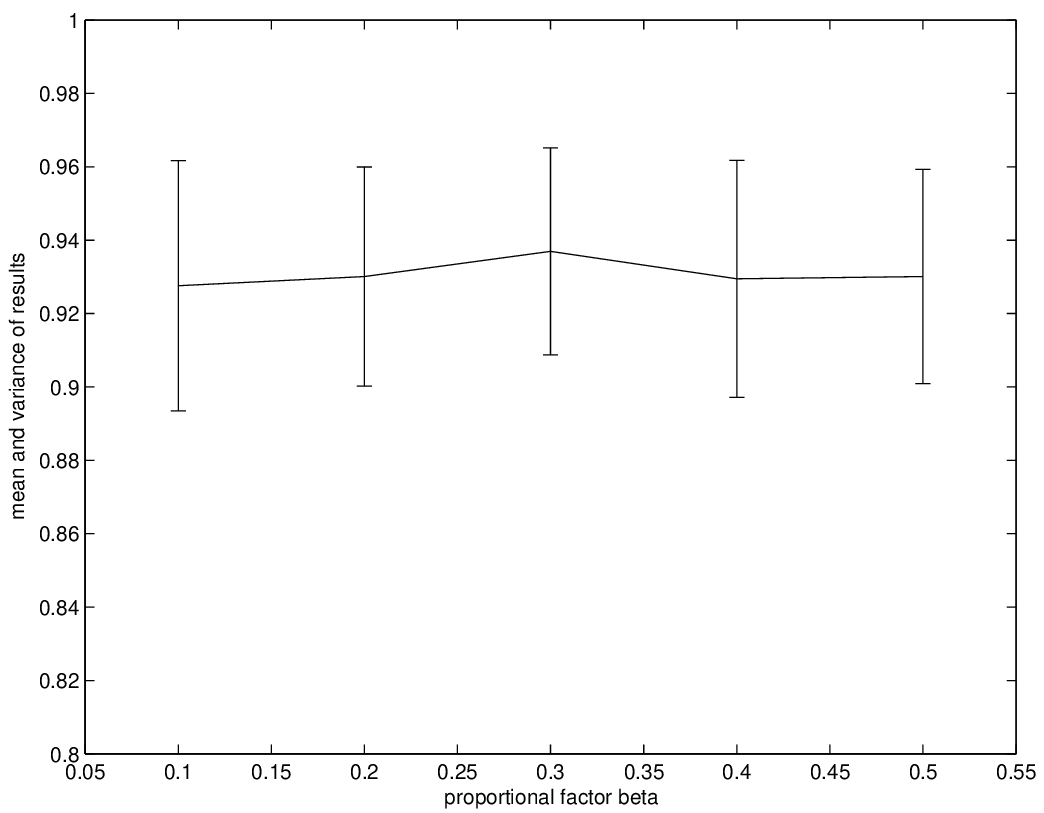}}\\
\caption{The effect of the cost for the clustering results. (a) the
results of the algorithm using the SD-like payoff matrix at
different number $k$ of nearest neighbors. (b) the mean and variance
corresponding to each curve in (a).}\label{fig:3}
\end{figure}

In the Snowdrift game, if the cost is too high, the SD-like payoff
matrix recovers the PD-like payoff matrix~\cite{Hauert2004}.
Therefore, the proportional factor $\beta$ is restricted in an
interval $(0, 0.5]$. However, different costs will bring about the
changes of the payoff matrix, which means that different clustering
results will be produced even in the same algorithm.
Figure~\ref{fig:3} illustrates how the clustering results change at
the different number \textit{k} of nearest neighbors when the
proportional factor $\beta$ takes different values, in which the
clustering results are represented by clustering accuracies. The
definition of clustering accuracy is given below.
\begin{definition}
$cluster_{i}$ is the label which is assigned to a data point
$\textbf{\textit{X}}_{i}$ in a dataset by the algorithm, and
$label_{i}$ is the actual label of the data point
$\textbf{\textit{X}}_{i}$ in the dataset. So the clustering accuracy
is \cite{Erkan2006}:
\begin{equation}
\begin{array}{ll}accuracy=\frac{\sum^{N}_{i=1}\lambda\big(map(cluster_{i}), label_{i}\big)}{N}\vspace{4pt}\\
\lambda(map(cluster_{i}), label_{i})=\left\{\begin{array}{lll}1 & \textrm{ if $map(cluster_{i})=label_{i}$}\vspace{4pt}\\
0 & \textrm{ otherwise}\end{array} \right.
\end{array}
\end{equation}
where the mapping function $map(\cdot)$ maps the label got by the
algorithm to the actual label.
\end{definition}
Clustering accuracy is an important evaluation criterion for
clustering algorithms, which reflects the level of matches between
the actual labels in a dataset and the labels assigned by a
clustering algorithm, so that the goal of a clustering algorithm is
to obtain higher clustering accuracies.

As is shown in Figure~\ref{fig:3}, it can be seen that similar
results are obtained by the algorithm at different costs, and the
best results are produced when $k=7$ and $k=8$, but from the
Figure~\ref{fig:3}(b), the clustering result with the largest mean
and the least variance is yielded when $\beta=0.3$. Also, as
mentioned above, the high cost leads to the recovery of the SD-like
payoff matrix, so the proportional factor $\beta=0.2$ or $\beta=0.3$
is recommended.

\subsection{Total payoffs and the rate of convergence} In this
subsection, at first, the total payoffs of algorithms using the PD-
or SD-like payoff matrix are compared respectively, and then the
differences between total payoffs are explained. Later, the rates of
convergence of algorithms are discussed when two LRR functions are
applied respectively, and further the impact for the rates of
convergence is analyzed in two payoff matrices.

\begin{figure}[htbp]
\centering \subfigure[]{ \label{fig4:subfig:a}
\includegraphics[width=0.45\textwidth]{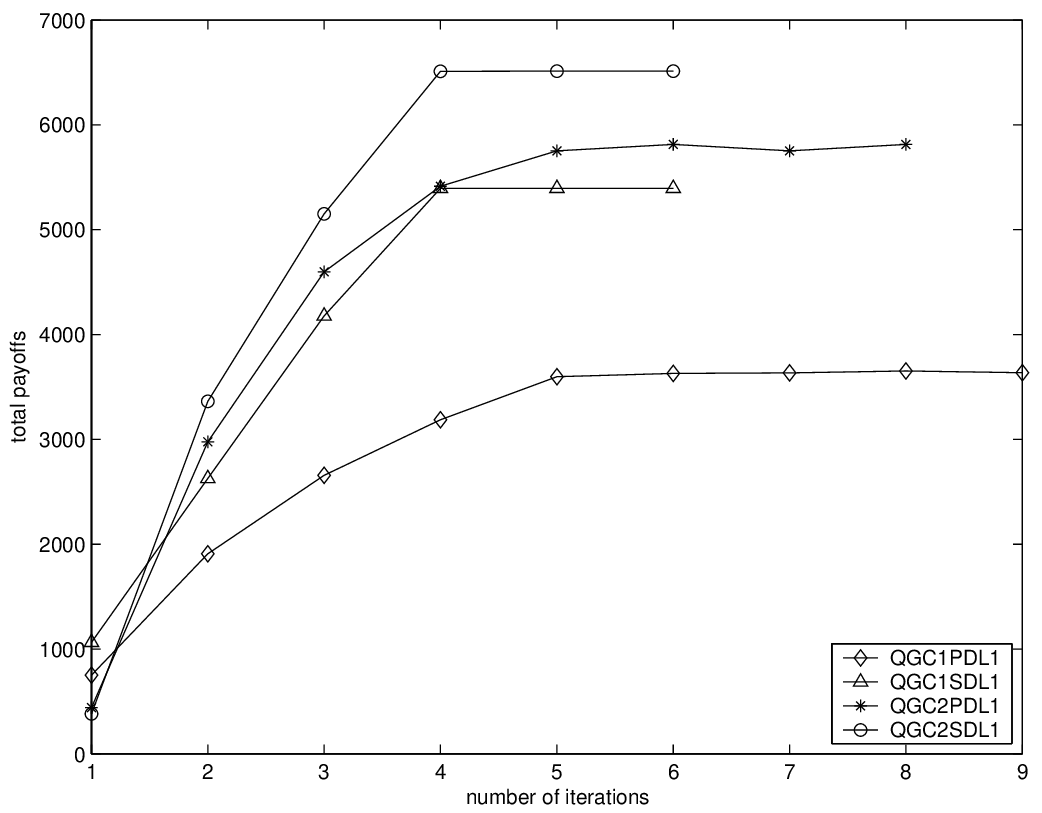}}
\subfigure[]{ \label{fig4:subfig:b}
\includegraphics[width=0.45\textwidth]{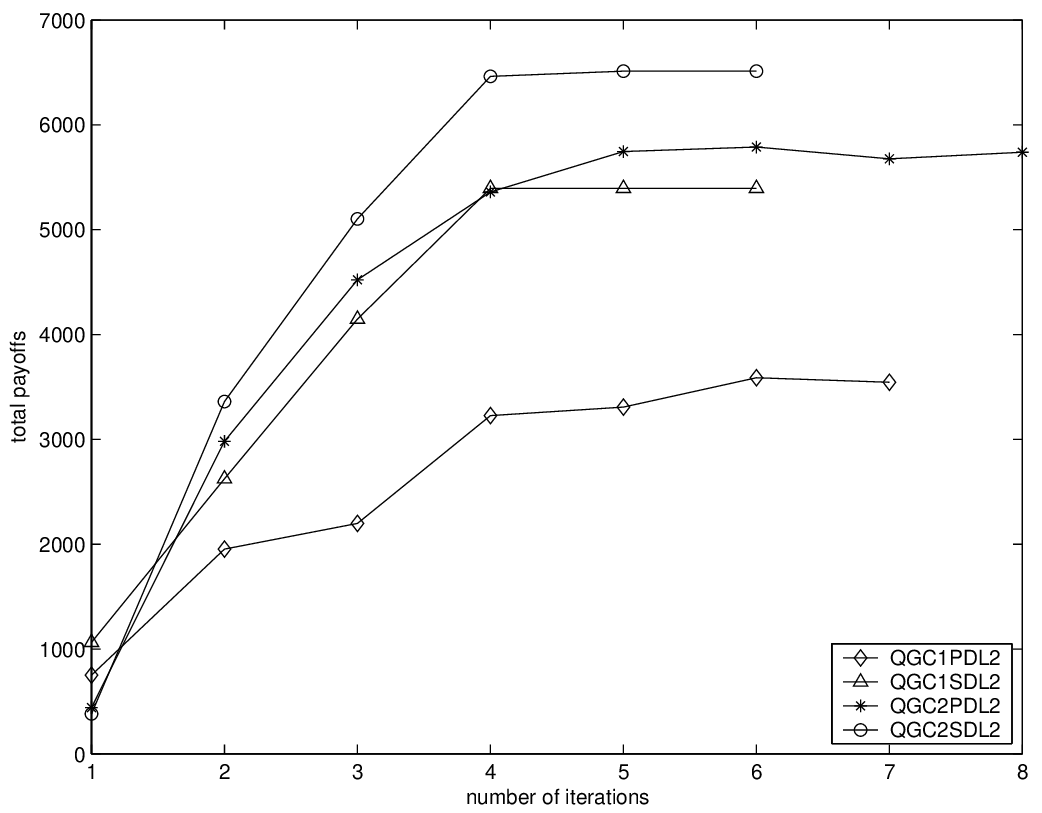}}\\
\caption{The total payoffs vs. the rates of convergence. (a) total
payoffs of algorithms in the case of LRR function
$L^{1}_{i}(\cdot)$, (b) total payoffs of algorithms in the case of
LRR function $L^{2}_{i}(\cdot)$}\label{fig:4}
\begin{remark}
The algorithms are named as follows. For example, the name of an
algorithm, QGC1PDL1, denotes that the Case 1, PD-like payoff matrix
and the LRR function $L^{1}_{i}(\cdot)$ are employed in this
algorithm.
\end{remark}
\end{figure}

If all other conditions are fixed, an algorithm will form two
versions due to using PD- or SD-like payoff matrix, and naturally
this will bring different results. As compared with the PD-like
payoff matrix, the payoff $P$ and $S$ in the SD-like payoff matrix
have a reverse order in the payoff inequality. In all algorithms,
the relationship between the total payoffs and the number of
iterations is drawn in Figure~\ref{fig:4}. From Figure~\ref{fig:4},
it can be found that the total payoffs in the algorithms with
SD-like payoff matrix are larger than that of the algorithms with
PD-like payoff matrix no matter which LRR function is selected.

According to two payoff matrices, using Eq.(\ref{eq:11}), each
player's expected payoff can be calculated in two cases of
strategies respectively as below.

Case 1:
\begin{equation}
\textrm{PD}:
z_{t}(\textbf{\textit{X}}_{i},\textbf{\textit{X}}_{j})=\left\{\begin{array}{ll}
\frac{1}{4}(0.6\omega+0.01\omega+\omega+0.2\omega)=0.4525\omega\vspace{4pt}\\
\frac{1}{2}(0.6\omega+0.01\omega)=0.305\omega
\end{array}\right.
\end{equation}
\begin{equation}
\textrm{SD}:
z_{t}(\textbf{\textit{X}}_{i},\textbf{\textit{X}}_{j})=\left\{\begin{array}{ll}
\frac{1}{4}(\omega-\frac{c}{2}+\omega-c+\omega+0.01\omega)=\frac{1}{4}(3.01-\frac{3\alpha}{2})\omega \vspace{4pt}\\
\frac{1}{2}(\omega-\frac{c}{2}+\omega-c)=\frac{1}{2}(2-\frac{3\alpha}{2})\omega
\end{array}\right.
\end{equation}

Case 2:
\begin{equation}
\textrm{PD}:
z_{t}(\textbf{\textit{X}}_{i},\textbf{\textit{X}}_{j})=\left\{\begin{array}{ll}
\rho_{1}\rho_{2}0.6\omega+\rho_{2}(1-\rho_{1})0.01\omega\\
+\rho_{1}(1-\rho_{2})\omega+(1-\rho_{1})(1-\rho_{2})0.2\omega\vspace{4pt}\\
(1-\rho_{1})0.6\omega+\rho_{1}0.01\omega
\end{array}\right.
\end{equation}
\begin{equation}
\textrm{SD}:
z_{t}(\textbf{\textit{X}}_{i},\textbf{\textit{X}}_{j})=\left\{\begin{array}{ll}
\rho_{1}\rho_{2}(\omega-\frac{c}{2})+\rho_{2}(1-\rho_{1})(\omega-c)\\
+\rho_{1}(1-\rho_{2})\omega+(1-\rho_{1})(1-\rho_{2})0.01\omega \vspace{4pt}\\
(1-\rho_{1})(\omega-\frac{c}{2})+\rho_{1}(\omega-c)
\end{array}\right.
\end{equation}

Comparing the expected payoffs in two cases, it can be observed that
when the SD-like payoff matrix is used, the expected payoff is
larger than that of using PD-like payoff matrix regardless of cases
of strategies. Therefore, this explains why differences between
total payoffs are produced.

Besides, Figure~\ref{fig:4} not only describes the changes of total
payoffs of algorithms, but also reflects the rates of convergence of
algorithms. When the total payoffs remain constant or fluctuate
slightly, this means that the algorithms have converged. At this
time, players do not frequently apply the LRR function to change his
neighbors but reach a stable state. As mentioned in the section
3.2, the LRR function $L^{2}_{i}(\cdot)$ only can observe an
extended neighbor set formed by those larger-than-average neighbors
in contrast to an extended neighbor set built by half neighbors in
the LRR function $L^{1}_{i}(\cdot)$. Generally speaking, for the
same \textit{k}, the median of payoffs is smaller than or equal to
the mean, i.e., $\theta^{1}_{t-1}\leq\theta^{2}_{t-1}$, which means
that the exploring area of the LRR function $L^{1}_{i}(\cdot)$ is
larger than that of the LRR function $L^{2}_{i}(\cdot)$. So, as a
whole, the algorithms with the LRR function $L^{2}_{i}(\cdot)$ are
slightly faster than that with the LRR function $L^{1}_{i}(\cdot)$,
i.e., the number of iterations that the former type of algorithms
needs is a little less.

Furthermore, in the algorithms, a phenomenon that the strategies
$\hat{H}$ and $\hat{F}_{t-1}$ are more likely used by players in a
high density area while the strategy $\hat{D}$ is used by those in a
low density area is always observed. This is because usually they
are mutually neighbors in the high density area on the weighted and
directed knn network, but in the low density area this case is
reverse. As a result, the differences of payoffs between the high
density and low density areas are enlarged rapidly for the expected
payoff in the strategy profile ($\hat{H}$,$\hat{H}$) or
($\hat{F}_{t-1}$,$\hat{F}_{t-1}$) is higher than that in other
strategy profile, and this also cause the distribution of players'
payoffs follows a power-law distribution, which is why algorithms
converge fast.

\section{Simulations}
To evaluate these clustering algorithms, six datasets are selected
from UCI repository \cite{Blake1998}, which are Soybean, Iris, Wine,
Sonar, Ionosphere and Breast cancer Wisconsin datasets, and complete
the simulations on them. In this section, firstly these datasets are
briefly introduced, and then the simulation results are
demonstrated.

The original data points in above datasets all are scattered in high
dimensional spaces spanned by their features which are the
individual measurable heuristic properties of the phenomena being
observed, where the description of all datasets is summarized in
Table~\ref{tab:4}. As for Breast dataset, some lost features are
replaced by random numbers, and the Wine dataset is standardized.
Finally, the algorithms are coded in Matlab 6.5.
\begin{table}[htbp]
\caption{Description of datasets.}\label{tab:4} \centering
{\begin{tabular} {cccc} \hline Dataset & Instances & Features &
classes
\\\hline
Soybean & \hphantom{0}47 & \hphantom{0}21 & 4 \\
Iris & \hphantom{0}150 & \hphantom{0}4 & 3 \\
Wine & \hphantom{0}178 & \hphantom{0}13 & 3 \\
Sonar & \hphantom{0}208 & \hphantom{0}60 & 2 \\
Ionosphere & \hphantom{0}351 & \hphantom{0}32 & 2\\
Breast & \hphantom{0}699 & \hphantom{0}9 & 2\\
\hline
\end{tabular}}
\end{table}

Throughout all simulations, data points in a dataset are viewed as
players in quantum games whose initial positions are taken from the
dataset. Next, the initial network representing relations among data
points are created according to Def.1, after a distance function is
selected, which only needs to satisfy that the more similar data
points are, the smaller the output of the function is. In the
simulations, the distance function is employed as following
\begin{equation}\label{eq:24}
d\big(\textbf{\textit{X}}_{i},\textbf{\textit{X}}_{j}\big)=exp\Big(\|\textbf{\textit{X}}_{i}-\textbf{\textit{X}}_{j}\|/2\sigma^{2}\Big),i,j=1,2,\cdots,N
\end{equation}
where the symbol $\|\cdot\|$ represents $L2$-norm. The advantage of
this function is that it not only satisfies our requirements, but
also overcomes the drawbacks of Euclidean distance. For instance,
when two points are very close, the output of Euclidean distance
function approaches zero, which may make the computation of payoff
fail due to the payoff approaching infinite. Nevertheless, when
Eq.(\ref{eq:24}) is selected as the distance function, it is more
convenient to compute the players' payoffs, since its minimum is one
and the reciprocals of its output are between zero and one,
$1/d(\textbf{\textit{X}}_{i},\textbf{\textit{X}}_{j})\in[0,1]$.

In addition, the distance between a player $\textbf{\textit{X}}_{i}$
and itself, $d(\textbf{\textit{X}}_{i},\textbf{\textit{X}}_{i})$, is
one according to the defined distance function, which means that
initially he is one of his \textit{k} nearest neighbors. So there is
an edge between the player $\textbf{\textit{X}}_{i}$ and himself,
namely a self-loop. Additionally, the parameter $\sigma$ in
Eq.(\ref{eq:24}) takes one. As is analyzed in the section 4.2, the
cost \textit{c} in SD-like payoff matrix is set by
$c=0.2\omega_{t}(\textbf{\textit{X}}_{i},\textbf{\textit{X}}_{j})$
in the related clustering algorithms.

The clustering algorithms are applied on the six datasets
respectively. Because two cases of strategies are designed and the
different payoff matrices and LRR functions are employed, the
algorithms are run on every dataset at the different number
\textit{k} of nearest neighbors. As is analyzed in section 4.1, for
a dataset the number \textit{k} of clusters decreases inversely with
the number of nearest neighbors. When a small \textit{k} is
selected, it is possible that the number of clusters is larger than
the preset number of the dataset, after the algorithm is ended. So a
merging-subroutine is called to merge unwanted clusters, which works
in this way. At first, the cluster with the fewest data points is
identified, and then it is merged to the cluster whose distance
between their centroids is smallest. This subroutine is repeated
till the number of clusters is equal to the preset number.

\begin{figure}[htbp]
\centering \subfigure[Soybean dataset]{
\label{fig5:subfig:a}\includegraphics[width=0.3\textwidth]{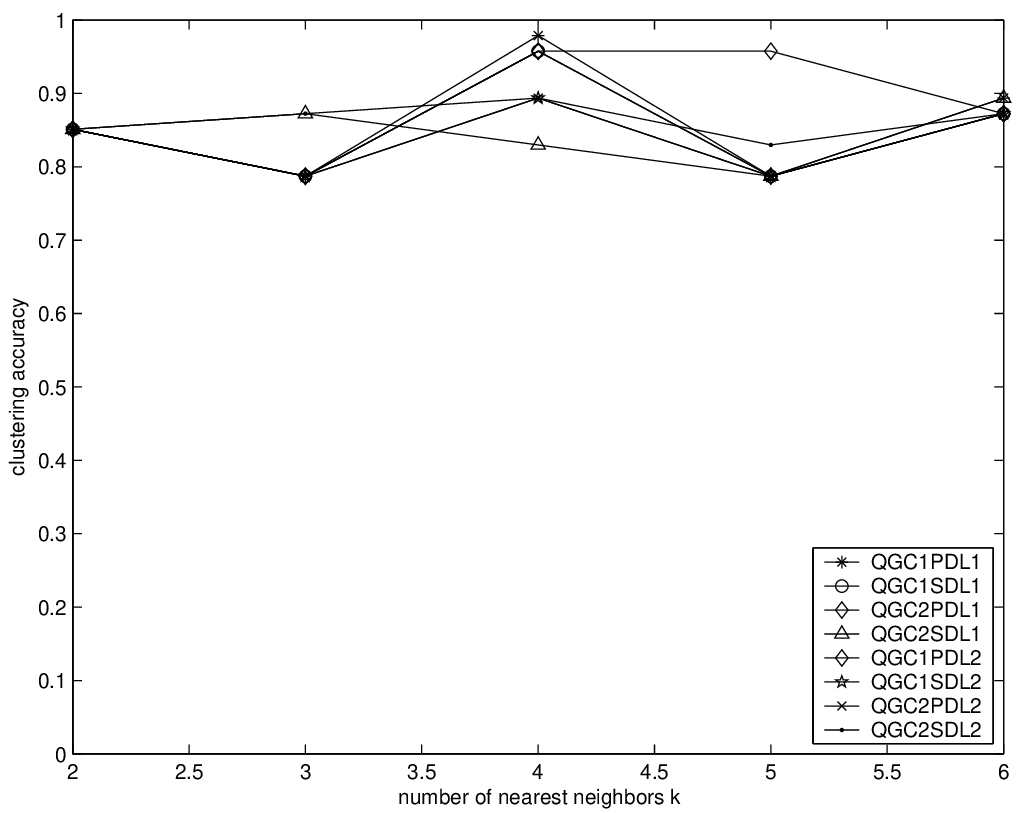}}
\subfigure[Iris dataset]{
\label{fig5:subfig:b}\includegraphics[width=0.3\textwidth]{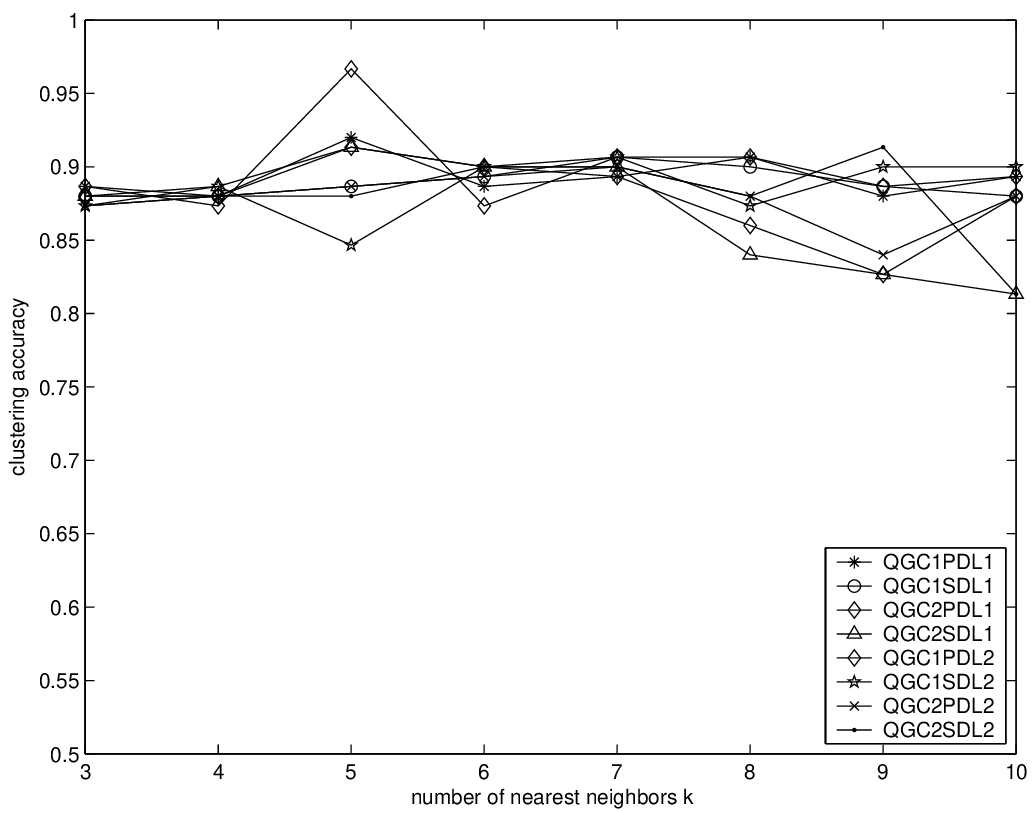}}
\subfigure[Wine dataset]{
\label{fig5:subfig:c}\includegraphics[width=0.3\textwidth]{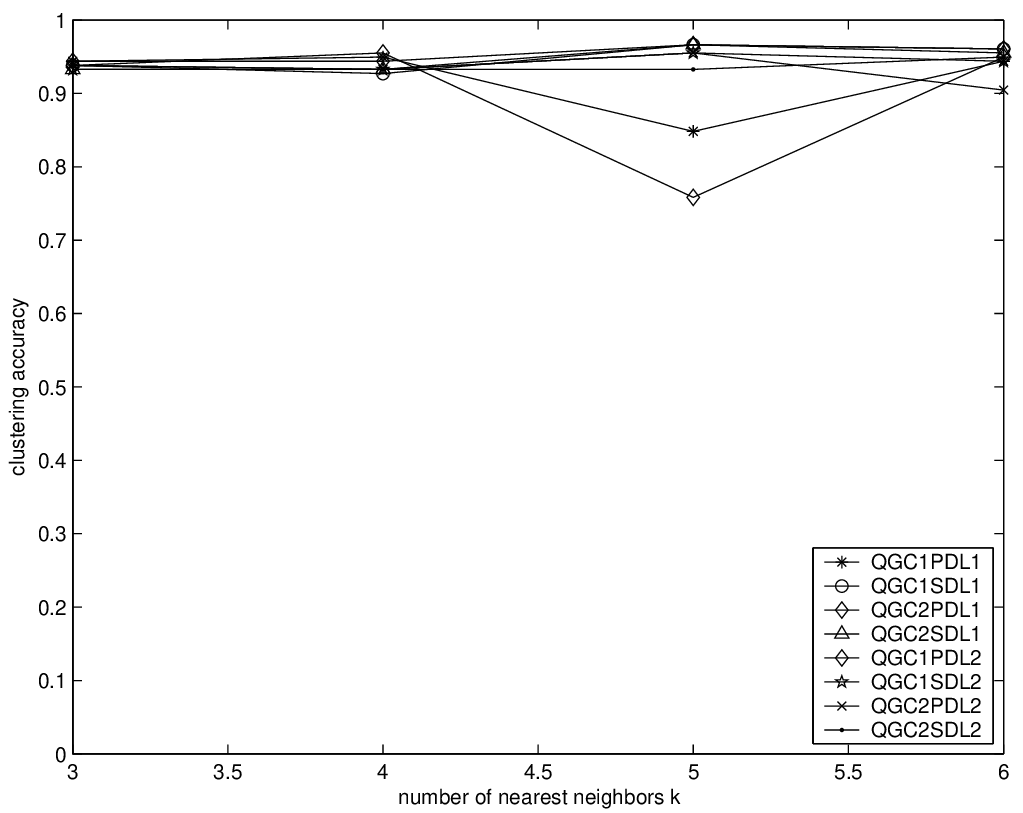}}\\
\subfigure[Sonar dataset]{
\label{fig5:subfig:d}\includegraphics[width=0.3\textwidth]{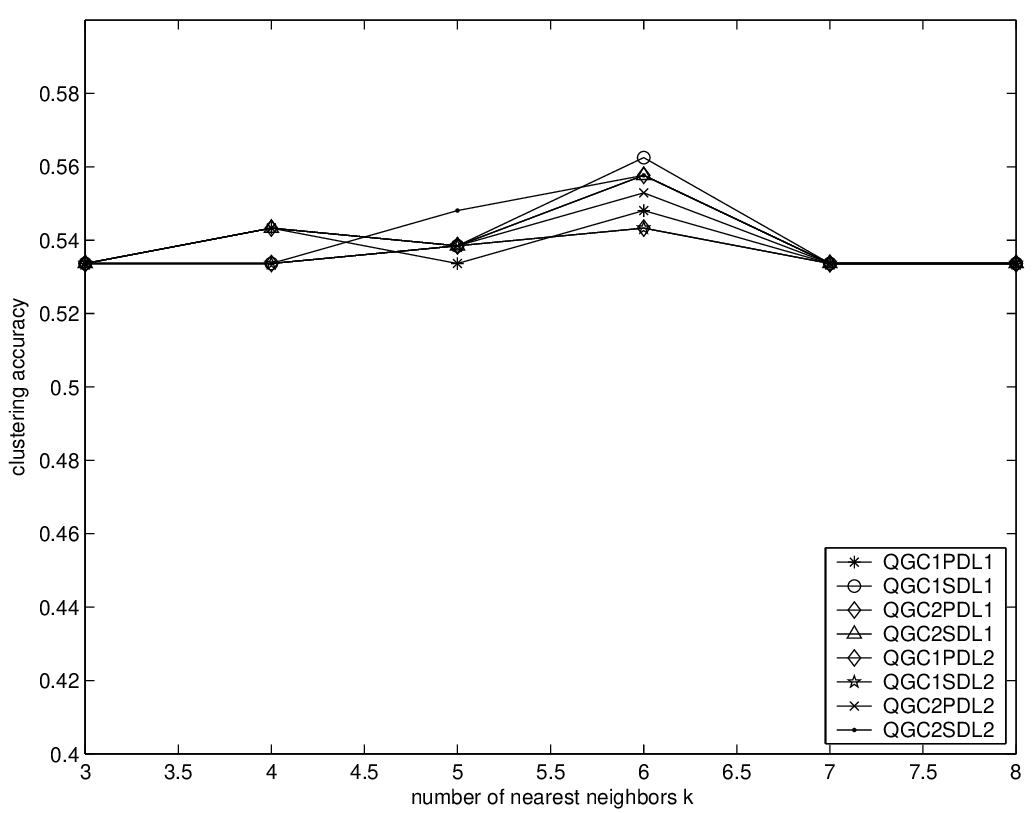}}
\subfigure[Ionosphere dataset]{
\label{fig5:subfig:e}\includegraphics[width=0.3\textwidth]{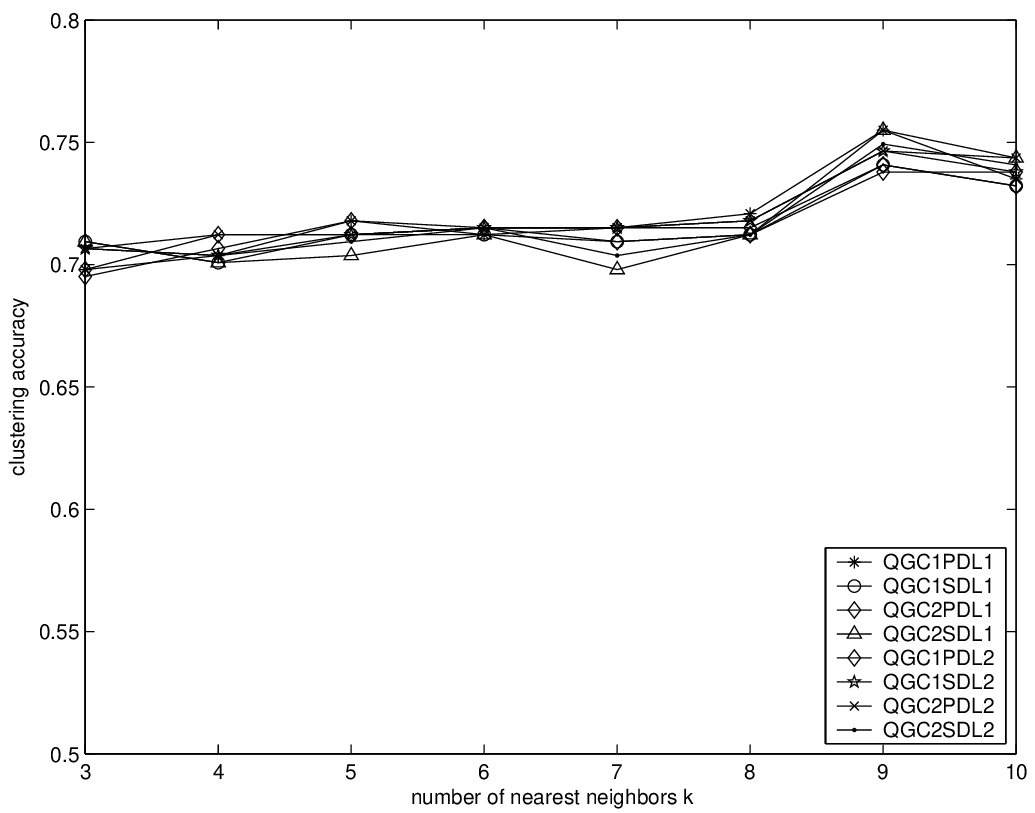}}
\subfigure[breast dataset]{
\label{fig5:subfig:f}\includegraphics[width=0.3\textwidth]{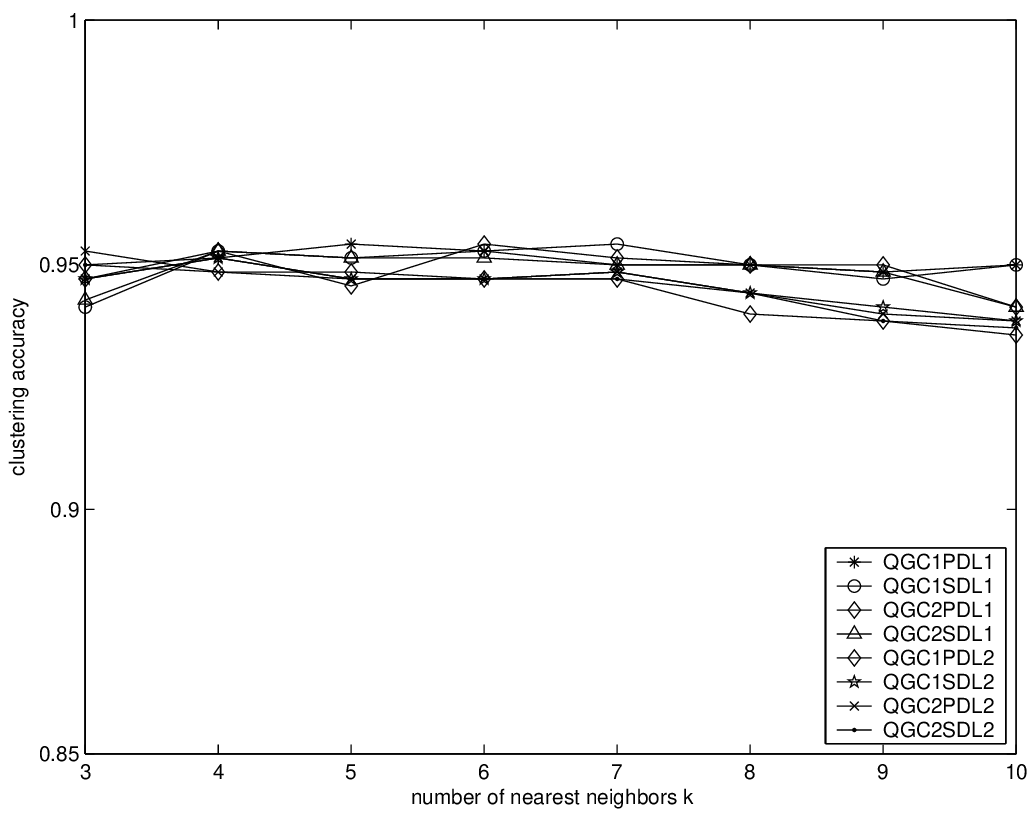}}\\
\caption{Comparison of clustering accuracies at different $k$ in all
proposed algorithms.}\label{fig:5}
\end{figure}

The clustering results obtained by these algorithms are compared in
Fig.~\ref{fig:5}, in which each point represents a clustering
accuracy. As is shown in Fig.~\ref{fig:5}, the similar results are
obtained by these algorithms at different number of nearest
neighbors, but almost all the best results are obtained by the
algorithms using the LRR function $L^{1}_{i}(\cdot)$.

Further, we show a simple comparison with three other algorithms:
Kmeans~\cite{MacQueen1967,Ding2007}, PCA-Kmeans \cite{Ding2007},
LDA-Km \cite{Ding2007}. The Kmeans algorithm is a popular clustering
algorithm because it is easy to implement. It begins with $k$
randomly-chosen cluster centers, and then assigns each data point in
a dataset to the closest cluster center. Next, the cluster centers
are recomputed according to the current cluster memberships. This
process will be repeated till a convergence criterion is met.
However, its major problem is sensitive to the selection of the
initial partition~\cite{Jain1999}. The PCA-Kmeans algorithm consists
of two steps: (1) Reduce the data dimension by principal component
analysis (PCA); (2) Clustering data points by the Kmeans algorithm.
Ding et al combine linear discriminant analysis (LDA) with the
Kmeans algorithm, and construct a new clustering algorithm called
'LDA-Km'. In the LDA-Km algorithm, Kmeans is used to generate class
labels, while LDA is used to do subspace selection. Finally, in the
subspace selection process, data points are clustered.

Our algorithms are established on the model of quantum games, so
data points in datasets are considered as players. This idea is
totally different from traditional clustering algorithms, such as
Kmeans and its variants, because in traditional clustering
algorithms data points for clustering are fixed, and various
functions or methods are designed to find cluster centers or
separating hyperplanes, whereas in the quantum-game-based clustering
algorithms data points (players) themselves choose their clusters,
which leads clusters are formed automatically during quantum games.
In conclusion, traditional clustering algorithms do it from outside,
while ours are from inside. Furthermore, our algorithms do not need
to choose cluster centers at beginning, so there does not exist the
problem that is "sensitive to the selection of the initial
partition". Besides, distances between data points are commonly used
as the measurement of their similarities in the Kmeans algorithm and
its variants. However, when similarities are measured in our
algorithms, not only distances between data points but also their
degrees and the strength of links are integrated, which provides
more information for the measurement of similarities. Later,
according to payoffs in quantum games, players apply a LRR function
to change the structure of the network, which makes the clusters
emerge. As a result, better clustering accuracies are obtained by
our algorithms.

Table~\ref{tab:5} displays the clustering accuracies of our
algorithms and the other three algorithms using the datasets in
Table~\ref{tab:4}. It can be observed that on almost all datasets
the quantum game based clustering algorithms have the best
clustering accuracies. Only on the Iris dataset, the LDA-Km
algorithm is better than ours, but our result is close to it as
well. The Iris dataset contains three clusters, one of which is far
from the other two, so data points in it can be clustered easily and
correctly. The other clusters in the Iris dataset, however, are
mixed partly in their boundaries, which brings a difficulty for
clustering. These mixed boundary points may be assigned to different
clusters by algorithms, which is why the clustering accuracies of
algorithms are various. It is more difficult to cluster data points
in the Sonar and Ionosphere dataset, because in these two datasets
plenty of data points belonging to different clusters are mixed
together. Therefore, the clustering accuracies of algorithms in them
are not high, and it is rather hard to improve the clustering
accuracies even by several percents. From the fifth and sixth
columns in Table~\ref{tab:5}, it can be seen that the quantum game
based clustering algorithms are better than other algorithms, which
indicates the effectiveness of our algorithms.
\begin{table}[htbp]\small
\caption{Comparison of clustering accuracies of
algorithms.}\label{tab:5} \centering {\begin{tabular} {ccccccc}
\hline Algorithm & Soybean & Iris & Wine & Sonar & Ionosphere &
Breast
\\\hline
QGC1PDL1 &\hphantom{0} 97.9\% &\hphantom{0} 92.0\% &\hphantom{0} 94.9\% &\hphantom{0} 54.8\% &\hphantom{0} 75.5\% &\hphantom{0} 95.4\% \\
QGC1SDL1 &\hphantom{0} 95.6\% &\hphantom{0} 90.7\% &\hphantom{0} 96.6\% &\hphantom{0} 56.3\% &\hphantom{0} 74.1\% &\hphantom{0} 95.4\% \\
QGC2PDL1 &\hphantom{0} 95.6\% &\hphantom{0} 91.3\% &\hphantom{0} 96.6\% &\hphantom{0} 55.8\% &\hphantom{0} 73.8\% &\hphantom{0} 95.4\% \\
QGC2SDL1 &\hphantom{0} 89.4\% &\hphantom{0} 91.3\% &\hphantom{0} 96.6\% &\hphantom{0} 55.8\% &\hphantom{0} 75.5\% &\hphantom{0} 95.3\% \\
QGC1PDL2 &\hphantom{0} 95.8\% &\hphantom{0} 96.7\% &\hphantom{0} 95.5\% &\hphantom{0} 54.3\% &\hphantom{0} 74.1\% &\hphantom{0} 95.0\% \\
QGC1SDL2 &\hphantom{0} 89.4\% &\hphantom{0} 90.7\% &\hphantom{0} 95.5\% &\hphantom{0} 54.3\% &\hphantom{0} 74.6\% &\hphantom{0} 95.1\% \\
QGC2PDL2 &\hphantom{0} 89.4\% &\hphantom{0} 90.0\% &\hphantom{0} 95.5\% &\hphantom{0} 55.3\% &\hphantom{0} 74.6\% &\hphantom{0} 95.3\% \\
QGC2SDL2 &\hphantom{0} 89.4\% &\hphantom{0} 91.3\% &\hphantom{0} 94.9\% &\hphantom{0} 55.8\% &\hphantom{0} 74.9\% &\hphantom{0} 95.1\% \\
Kmeans~\cite{Ding2007} & \hphantom{0}68.1\% & \hphantom{0}89.3\% & \hphantom{0}70.2\% & \hphantom{0}47.2\% & \hphantom{0}71.0\% & \hphantom{0}-- \\
PCA-Kmeans~\cite{Ding2007} & \hphantom{0}72.3\% & \hphantom{0}88.7\% & \hphantom{0}70.2\% & \hphantom{0}45.3\% & \hphantom{0}71.0\% & \hphantom{0}-- \\
LDA-Km~\cite{Ding2007} & \hphantom{0}76.6\% & \hphantom{0}98.0\% & \hphantom{0}82.6\% & \hphantom{0}51.0\% & \hphantom{0}71.2\% & \hphantom{0}-- \\
\hline
\end{tabular}}
\end{table}

\section{Conclusions}
The enormous successes gained by the quantum algorithms make us
realize it is possible that the quantum algorithms can obtain
solutions faster and better than those classical algorithms for some
problems. Therefore, we combine the quantum game with the problem of
data clustering, and establish clustering algorithms based on
quantum games. In the algorithms, data points for clustering are
regarded as players who can make decisions in quantum games. On a
weighted and directed knn network that represents relations among
players, each player uses quantum strategies against every one of
his neighbors in a $2\times2$ entangled quantum game respectively.
We design two cases of strategies: (i) one plays the strategy
$\hat{H}$, the other plays the strategy $\hat{H}$ or $\hat{D}$, (ii)
one plays the strategy $\hat{F}_{t-1}$, the other plays the strategy
$\hat{F}_{t-1}$ or $\hat{D}$ according to the strength of links, in
each of which players' expected payoffs are calculated based on the
PD- and SD-like payoff matrices respectively. According to
neighbors' payoffs in one's neighbor set, each player applies a LRR
function ($L^{1}_{i}(\cdot)$ or $L^{2}_{i}(\cdot)$) to change his
neighbors, i.e., the links connecting to neighbors with small
payoffs are removed and then new links are created to those with
higher payoffs. Later, the Grover iteration \textit{G} is employed
to update the strength of links between him and his neighbors. In
the process of playing quantum game, the structure of the network
formed by players tends to stability gradually. In other words, each
player always connects to one of his neighbors with the largest
strength or jumps among some neighbors with the largest strength in
a constant period. At this time, if only the links with the largest
strength are left but all other links are removed among players, the
network naturally divides into several separate parts, each of which
corresponds to a cluster.

Additionally, in simulations, it can be found that the total
expected payoffs of algorithms using SD-like payoff matrix are
higher than that of algorithms using PD-like payoff matrix. Later,
the reason is explained. Further, we observe that the rates of
convergence of the algorithms employing the LRR function
$L^{2}_{i}(\cdot)$ are faster slightly than that of the algorithms
employing the LRR function $L^{1}_{i}(\cdot)$, because more areas
are explored by the LRR function $L^{1}_{i}(\cdot)$, but this brings
better clustering results. In the case when the exact number of
clusters is unknown in advance, one can adjust the number \textit{k}
of nearest neighbors to control the number of clusters, where the
number of clusters decreases inversely with the number \textit{k} of
nearest neighbors. Finally, the clustering algorithms are evaluated
on six real datasets, and simulation results have demonstrated that
data points in a dataset are clustered reasonably and efficiently.

\section*{Acknowledgments}
The authors would like to thank the anonymous referees for their
helpful comments and suggestions to improve the presentation of this
paper. This work was supported in part by the National Natural
Science Foundation of China (No. 60405012, No. 60675055).

\bibliography{Manuscript}   
\bibliographystyle{hieeetr}

\end{document}